\documentclass[tikz]{article}

\usepackage[nonatbib,dandb, final]{neurips_2025}

\usepackage[utf8]{inputenc} % allow utf-8 input
\usepackage[T1]{fontenc}    % use 8-bit T1 fonts
\usepackage{hyperref}       % hyperlinks
\usepackage{url}            % simple URL typesetting
\usepackage{booktabs}       % professional-quality tables
\usepackage{longtable}
\usepackage{amsfonts}       % blackboard math symbols
\usepackage{nicefrac}       % compact symbols for 1/2, etc.
\usepackage{microtype}      % microtypography
\usepackage{xcolor}         % colors
\usepackage{colortbl} % For adding colour to tables
\usepackage{hhline}
\usepackage{graphicx}
\usepackage{fontawesome}
\usepackage{subfig}
\usepackage{placeins}
\usepackage[most]{tcolorbox}
\tcbuselibrary{skins}
\tcbuselibrary{breakable}
\usepackage{enumitem}
\usepackage{tabularx}
\usepackage{booktabs}
\usepackage{amsmath}
\colorlet{linecol}{black!75}
\usepackage{forest}
\tikzset{
  my rounded corners/.append style={rounded corners=2pt},
}

\definecolor{customdefaultcolor}{HTML}{7A899F}
\definecolor{customblue}{HTML}{1E90FF}

\definecolor{checklistbox-color}{HTML}{D14D41}
\definecolor{checklistbox-color-bg}{HTML}{FFE1D5}

\definecolor{examplebox-color}{HTML}{4874B6}
\definecolor{examplebox-color-bg}{HTML}{D3E5F5}

\definecolor{phenomenon}{HTML}{4285be}
\definecolor{claims}{HTML}{879939}
\definecolor{task}{HTML}{da702c}
\definecolor{metric}{HTML}{e47da8}

\usepackage{svg}

\tcbset{
  base/.style={
    empty,
    frame engine=path,
    skin first = enhanced,
    skin middle = enhanced,
    skin last = enhanced,
    colframe=checklistbox-color-bg,
    interior style={fill=checklistbox-color-bg},
    sharp corners,
    attach boxed title to top left={yshift*=-\tcboxedtitleheight},
    boxed title style={
        size=minimal,
        top=4pt,
        left=4pt,
        overlay={
            \draw[checklistbox-color,line width=3pt,] (frame.north west)--(frame.north east);
        }
    },
    coltitle=checklistbox-color,
    colback=checklistbox-color,
    fonttitle=\mdseries,
    overlay unbroken and first={
        \draw[checklistbox-color](title.north east)--([xshift=-0.5pt]title.north east-|frame.east);
    },
    breakable
}}
\tcbset{
  example/.style={
    empty,
    frame engine=path,
    skin first = enhanced,
    skin middle = enhanced,
    skin last = enhanced,
    colframe=examplebox-color-bg,
    interior style={fill=examplebox-color-bg},
    sharp corners,
    attach boxed title to top left={yshift*=-\tcboxedtitleheight},
    boxed title style={
        size=minimal,
        top=4pt,
        left=4pt,
        overlay={
            \draw[examplebox-color,line width=3pt,] (frame.north west)--(frame.north east);
        }
    },
    coltitle=examplebox-color,
    colback=examplebox-color,
    fonttitle=\mdseries,
    overlay unbroken and first={
        \draw[examplebox-color](title.north east)--([xshift=-0.5pt]title.north east-|frame.east);
    },
    breakable
}}
\newtcolorbox{checklistbox}[2][]{%
  base,
  title={\hspace{4pt}\includegraphics[scale=.6]{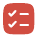}\hspace{3.5pt}#2},
}
\newtcolorbox{examplebox}[2][]{%
  example,
  title={\hspace{4pt}\hspace{3.5pt}\Large{#2}},
}

\usepackage[maxbibnames=5,maxcitenames=2, sorting=none,doi=false,url=false,eprint=false,giveninits=true]{biblatex}
\AtEveryBibitem{%
\ifentrytype{article}{
    \clearfield{issn}
    \clearfield{url}%
    \clearfield{urlyear}%
    \clearfield{month}%
    \clearfield{pages}%
    \clearfield{volume}%
    \clearfield{number}%
}{}

\ifentrytype{inproceedings}{
    \clearfield{url}%
    \clearfield{urlyear}%
    \clearfield{month}
    \clearfield{address}%
    \clearfield{pages}%
    \clearfield{isbn}
}{}

\ifentrytype{misc}{
    \clearfield{primaryclass}%
    \clearfield{url}%
    \clearfield{issn}%
    \clearfield{urlyear}%
    \clearfield{month}
    \clearfield{journal}%
    \clearfield{doi}%
}{}
}

\DeclareSourcemap{
  \maps[datatype=bibtex]{
    \map{
      \step[fieldset=primaryclass,null]
    }
  }
}

\usepackage{xspace}

\definecolor{myred}{RGB}{255,138,103}

\usepackage{enumitem}

\usepackage[nameinlink,capitalise]{cleveref} % For intelligent cross-referencing
\crefname{section}{\S}{\S} % Custom section referencing
\crefname{Section}{\S}{\S} % Custom Section referencing
\crefformat{appendix}{#2App.~#1#3} % Custom appendix referencing
\crefname{appendix}{App.}{Apps.}
\crefname{table}{Tab.}{Tab.}
\crefname{appendix_table}{Tab.}{Tab.}
\crefname{Table}{Tab.}{Tab.}
\crefname{Figure}{Fig.}{Fig.}
\crefname{figure}{Fig.}{Fig.}

\usepackage[toc,page,header]{appendix}
\usepackage{minitoc}
\title{Measuring what Matters: Construct Validity in Large Language Model Benchmarks}

{\centering
\author{\textbf{Andrew M. Bean}$^1$\thanks{andrew.bean@oii.ox.ac.uk,\quad adam.mahdi@oii.ox.ac.uk}\quad\textbf{Ryan Othniel Kearns}$^{1}$\quad\textbf{Angelika Romanou}$^2$\\\textbf{Franziska Sofia Hafner}$^1$\quad\textbf{Harry Mayne}$^1$\\ \\\textbf{Jan Batzner}$^{3,4}$ \quad \textbf{Negar Foroutan}$^2$\quad\textbf{Chris Schmitz}$^5$\quad\textbf{Karolina Korgul}$^1$\quad \textbf{Hunar Batra}$^1$\\\textbf{Oishi Deb}$^1$\quad\textbf{Emma Beharry}$^6$\quad\textbf{Cornelius Emde}$^1$\quad\textbf{Thomas Foster}$^1$\quad\textbf{Anna Gausen}$^7$\\\textbf{Mar\'ia Grandury}$^{8,9}$\quad\textbf{Simeng Han}$^{10}$\quad\textbf{Valentin Hofmann}$^{11,12}$\quad\textbf{Lujain Ibrahim}$^1$\\\textbf{Hazel Kim}$^1$\quad\textbf{Hannah Rose Kirk}$^{1,7}$\quad\textbf{Fangru Lin}$^1$\\\textbf{Gabrielle Kaili-May Liu}$^{10}$\quad\textbf{Lennart Luettgau}$^7$\quad\textbf{Jabez Magomere}$^1$\quad\textbf{Jonathan Rystr{\o}m}$^1$\\\textbf{Anna Sotnikova}$^2$\quad\textbf{Yushi Yang}$^1$\quad\textbf{Yilun Zhao}$^{10}$\\ \\\textbf{Adel Bibi}$^1$\quad\textbf{Antoine Bosselut}$^2$\quad\textbf{Ronald Clark}$^1$\quad\textbf{Arman Cohan}$^{10}$\quad\textbf{Jakob Foerster}$^1$\\\textbf{Yarin Gal}$^{1,7}$\quad\textbf{Scott A. Hale}$^{1,13}$\quad\textbf{Inioluwa Deborah Raji}$^{14}$\quad\textbf{Christopher Summerfield}$^{1,7}$\\\textbf{Philip H.S. Torr}$^1$\quad\textbf{Cozmin Ududec}$^7$\quad\textbf{Luc Rocher}$^1$\quad\textbf{Adam Mahdi}$^{1*}$\\ \\$^1$University of Oxford \quad $^2$EPFL\quad$^3$ Weizenbaum Institute Berlin\\ $^4$Technical University Munich\quad$^5$Centre for Digital Governance, Hertie School\\$^6$Stanford University\quad$^7$UK AI Security Institute\quad$^8$SomosNLP\\$^9$Universdad Polit\'ecnica de Madrid\quad$^{10}$Yale University\quad$^{11}$Allen Institute for AI\\$^{12}$University of Washington\quad$^{13}$Meedan\quad$^{14}$UC Berkeley
}}

\begin{document}

\maketitle

\doparttoc % Tell to minitoc to generate a toc for the parts
% Adding TOC now for understanding our structure
\faketableofcontents % Run a fake tableofcontents command for the partocs

\begin{abstract}
   Evaluating large language models (LLMs) is crucial for both assessing their capabilities and identifying safety or robustness issues prior to deployment. Reliably measuring abstract and complex phenomena such as `safety' and `robustness' requires strong \textit{construct validity}, that is, having measures that represent what matters to the phenomenon. With a team of 29 expert reviewers, we conduct a systematic review of 445 LLM benchmarks from leading conferences in natural language processing and machine learning. Across the reviewed articles, we find patterns related to the measured phenomena, tasks, and scoring metrics which undermine the validity of the resulting claims. To address these shortcomings, we provide eight key recommendations and detailed actionable guidance to researchers and practitioners in developing LLM benchmarks.
\end{abstract}

%%%%%%% SECTIONS %%%%%
\begin{refsection}
\section{Introduction}
Benchmarks and evaluations play a critical role in the development of large language models. They help determine which model improvements are considered useful and set the direction of future research~\cite{dotanValueladenDisciplinaryShifts2019,hutchinsonEvaluationGapsMachine2022}. Creating a benchmark requires operationalising phenomena (abstract concepts) into concrete tasks and metrics that serve as measurable proxies for model capabilities~\cite{rajiAIEverythingWhole2021a}. As an example, the `intelligence' of LLMs is frequently debated~\cite{bubeckSparksArtificialGeneral2023, mccoyEmbersAutoregressionUnderstanding2023b}, but cannot be measured directly, making it necessary to develop proxies~\cite{hernandez-oralloMeasureAllMinds2017}. The value of a benchmark depends on whether it is a good proxy for the real-world phenomenon it intends to measure. This property is known as \textit{construct validity}: the degree to which a benchmark score provides evidence for making claims about the target phenomenon~\cite{bowmanWhatWillIt2021,rajiAIEverythingWhole2021a,cronbachConstructValidityPsychological1955}. If a benchmark has high construct validity in measuring `intelligence', then a model which does well is in some sense `intelligent', but if the construct validity is low, then a high score may be irrelevant or even misleading.

The science of evaluating large language models (LLMs) is still in its early stages, with a pressing need for shared standards and best practices~\cite{weidingerEvaluationScienceGenerative2025, bowmanWhatWillIt2021}. Certain specific issues such as reproducibility and cost have been addressed via shared implementation standards~\cite{bidermanLessonsTrenchesReproducible2024a,UK_AI_Security_Institute_Inspect_AI_Framework_2024}, and item selection methods~\cite{poloTinyBenchmarksEvaluatingLLMs2024a}, respectively. Other issues, such as the best use of statistical methods~\cite{mcintoshInadequaciesLargeLanguage2024, millerAddingErrorBars2024, luettgauHiBayESHierarchicalBayesian2025} and social responsibility~\cite{bowmanWhatWillIt2021, gebruDatasheetsDatasets2021} have also been raised. \textcite{reuelBetterBenchAssessingAI2024} aggregate best practices and provide recommendations for the whole lifecycle of a benchmark. However, identifying concrete best practices for creating benchmarks with high construct validity remains a difficult task. Benchmarks with low construct validity have real consequences, since unrecognised weak links between tasks and the underlying phenomena they claim to measure can lead to poorly supported scientific claims, misdirected research, and policy implications that are not grounded in robust evidence.

Here, we assess practices around the construct validity of LLM benchmarks through a systematic review of 445 articles from leading ML and NLP conferences. The articles were coded by experts in ML and NLP using a detailed conceptual and methodological schema that identifies useful practices in the design and interpretation of benchmarks for increasing the validity of measurements. Almost all articles have weaknesses in at least one area across phenomena, tasks, metrics, and claims. Key concepts are often poorly defined or operationalised, limiting the reliability of the conclusions they draw. We call for improved practices and reporting standards for establishing construct validity in new benchmarks, and release an operational checklist of best practice recommendations.

\begin{figure}
    \centering
    \includegraphics[width=\linewidth]{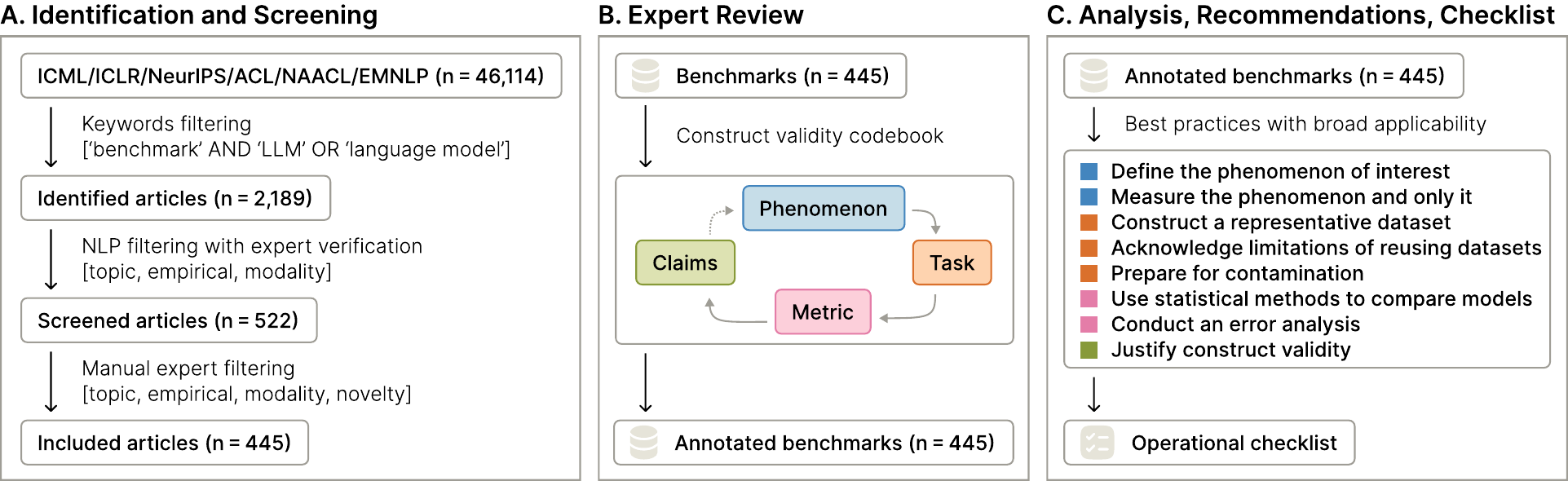}
    \caption{\textbf{Systematic review process.} (A) Identification and screening from relevant proceedings. (B) In-depth review and annotation of included benchmarks. A phenomenon is operationalised via a task, scored with a metric, to support a claim about this phenomenon. (C) Synthesis of best practices.}
    \label{fig:systematic-review-process}
\end{figure}

\section{Background}\label{background}

Construct validity evaluates whether an empirical test measures the phenomenon it intends to measure~\cite{alaaMedicalLargeLanguage2025a}. Formal assessments of construct validity originate from psychological testing as a means of creating tests for phenomena which cannot be directly verified, such as personality~\cite{cronbachConstructValidityPsychological1955}.

Construct validity as an overarching concept can be assessed by considering  various features of the test design~\cite{messickTestValidityMatter1998}. At the level of phenomena, \textit{face validity} considers whether a test appears prima facie a valid representation of the phenomenon~\cite{nevoFaceValidityRevisited1985,mosierCriticalExaminationConcepts1947}. At the task level, \textit{content validity} considers whether the task content represents all important aspects of the phenomenon being measured~\cite{sireciConstructContentValidity1998}. \textit{Ecological} and \textit{predictive validity} concern the relevance of the test to real-world settings~\cite{schmucklerWhatEcologicalValidity2001}, including how it predicts future performance~\cite{alaaMedicalLargeLanguage2025a}. \textit{Convergent}, \textit{discriminant}, and \textit{criterion validity} measure whether test findings correlate with, and only with, tests for similar phenomena~\cite{campbellConvergentDiscriminantValidation1959}.

With LLMs, construct validity is key for benchmarking abstract abilities such as `reasoning.' The value of construct validity has been emphasised in previous NLP literature~\cite{bowmanWhatWillIt2021, rajiAIEverythingWhole2021a}. Standard benchmarks and narrowly-defined tasks are now quickly becoming saturated~\cite{kielaDynabenchRethinkingBenchmarking2021d} and attention is shifting towards testing general-purpose abilities of LLMs~\cite{wangSuperGLUEStickierBenchmark2019,singhGlobalMMLUUnderstanding2025}. The interpretation of such evaluations has become contested, with disagreements about whether results show signs of intelligence~\cite{bubeckSparksArtificialGeneral2023, mccoyEmbersAutoregressionUnderstanding2023b} or emergent abilities~\cite{schaefferAreEmergentAbilities2023a,  weiEmergentAbilitiesLarge2022}, making assessing construct validity all the more crucial.
\section{Methods}

\paragraph{Study design}
\label{sec:selecting}
We conducted a systematic review, as illustrated in  \Cref{fig:systematic-review-process}. Our corpus consisted of 46,114 articles drawn from the proceedings of ICML, ICLR and NeurIPS (accessed via proceedings websites)  between 2018 and 2024, and from ACL, NAACL and EMNLP between 2020 and 2024 (accessed via ACL Anthology). The ACL range was limited by abstract availability.

We identified and selected articles whose titles or abstracts contained the keywords `benchmark' and either `LLM' or `language model', resulting in an initial set of 2,189 articles, with most articles coming from recent years, and only 14 in 2018 and 2019.

We applied four inclusion criteria to assess the relevance and suitability of each article. First, we evaluated whether the article concerned the capabilities of LLMs, excluding those focused solely on technical aspects such as inference speed or energy consumption. Second, we determined whether the article introduced an empirical benchmark and reported LLM performance, excluding opinions, reviews or policy frameworks. Third, we assessed whether the benchmark was compatible with text and vision models, filtering out those that required other modalities such as audio or video. Finally, we checked that the article introduced a novel benchmark or made a substantial modification to an existing one, excluding repackaged or minimally altered combinations of prior benchmarks. 

%This criterion excludes articles that focus only on technical performance, such as computational efficiency or inference speed.

%     \item \textit{Empirical}: \textit{Does the article implement a benchmark and report the performance of large language models?} This criterion excludes opinions, reviews or policy frameworks.

%     \item \textit{Modality}: \textit{Could a text and vision model be tested on the benchmark?} This criterion excludes articles requiring video, audio, or other multimodal abilities that are not the focus of our review. 

%     \item \textit{Novelty}: \textit{Does the article introduce a new benchmark or meaningfully alter an existing one?} This criterion excludes articles which combine or repackage existing benchmarks with only minimal changes.
%\end{enumerate}

We first used GPT-4o mini \cite{gpt-4o-mini} to screen the articles on the basis of the first three criteria.  This model-assisted step was validated against human-labelled data for  a sample of 50 articles and achieved an F1 score of 84\%. This automated step reduced the set to 522 articles eligible for manual filtering. We then assigned the 522 eligible articles to 29 reviewers matched on area of expertise, to manually determine inclusion using all four criteria, resulting in 445 articles included for final review.

\paragraph{Codebook and expert review}\label{codebook}

We created an initial a priori codebook for phenomena, tasks, metrics and claims. Building on the definition of a benchmark from \textcite{rajiAIEverythingWhole2021a}, we consider a benchmark to be a `task' and `metric' which are used together to represent a `phenomenon' of interest. These elements are considered alongside the interpretation of the results by the authors.\footnote{We chose to use the terms `task' and `phenomenon', rather than `dataset' and `task', to better capture benchmarks that involve dynamic elements or aim to measure abstract capabilities rather than specific tasks.} For example, in GSM8K \cite{cobbe_training_2021}, the \textit{phenomenon} is `multi-step mathematical reasoning', which is measured via the \textit{task} of answering short free response questions drawn from grade-school mathematics word problems, which are scored via the `exact match' \textit{metric}.

Items in the codebook were derived deductively based on prior literature to provide indications of key aspects of construct validity, including face, predictive, content, ecological, convergent and discriminant validity (see~\Cref{background}). Each article was coded by a primary reviewer using this codebook. A second reviewer mapped the responses onto a simplified list of options for computing statistics and these mappings were verified by the primary reviewer. A random sample of 46 papers were reviewed twice, with a mean Brennan–Prediger Kappa of .524 across all 30 categorical questions.
%
%\paragraph{Analysis, Recommendations and Checklist} 
The first author then read a subset of 50 articles and reviewed all the 445 annotations to synthesise the findings into an initial set of recommendations through an inductive open coding process. Finally, these recommendations were collaboratively refined through an iterative process involving multiple authors across five meetings.

\section{Results}
\label{sec:taxonomies}

\begin{figure}[t!]
    \centering
    \includegraphics[width=\linewidth]{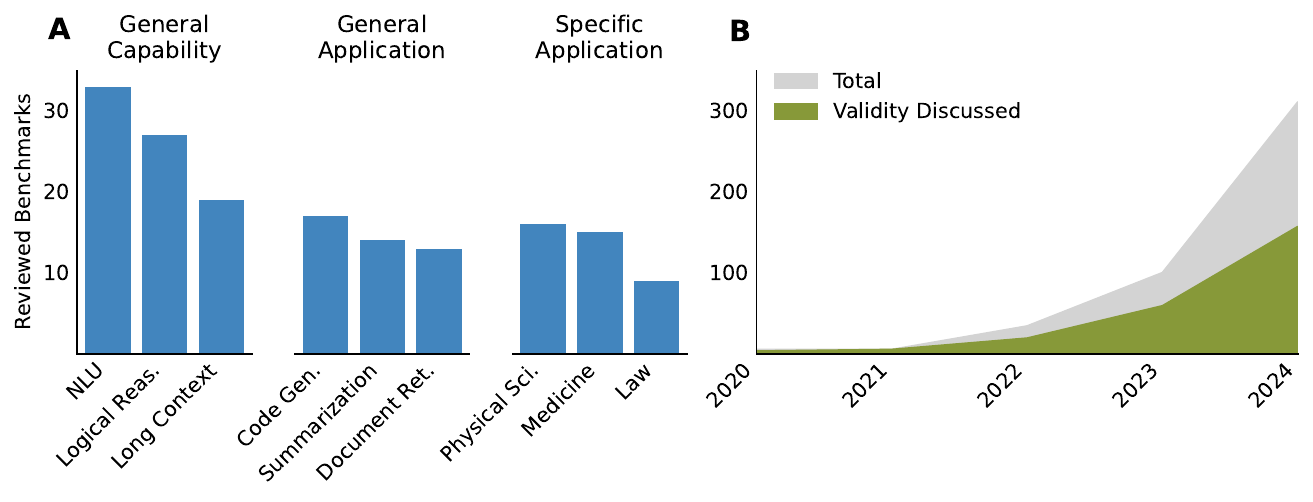}
    \caption{\textbf{Summary of reviewed articles.} (A) Three most common categories of benchmark phenomena, grouped into general capabilities, general applications, and specific applications. (B) Number of articles by publication year and number which discuss the construct validity of their benchmark.}
    \label{fig:results_overview}
\end{figure}

\begin{figure}
    \centering
    \includegraphics[width=\linewidth]{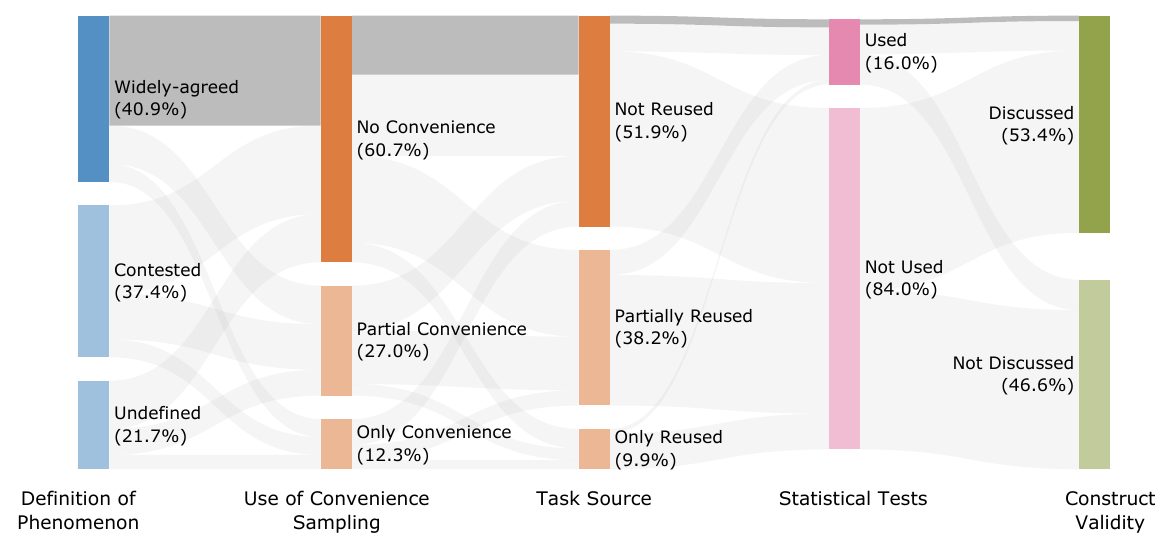}
    \caption{\textbf{Key codebook results.} The distribution of codebook responses on selected items. In each column, the options are ordered from most to least preferred for high construct validity. The shaded area indicates the benchmarks that follow the best practices for all five items.}
    \label{fig:results_sankey}
\end{figure}

The reviewing process resulted in a dataset containing responses to 21 question items on 445 benchmark articles, annotated by 29 experts in the areas of NLP and machine learning. \Cref{fig:results_overview} shows that the number of included articles increases significantly in each subsequent year. The dataset contains information covering all of the stages of the benchmarks, from how they initially define their phenomenon of interest, to which tasks they select in an attempt to measure this phenomenon, to the metrics they use to estimate and compare the performance of language models on these tasks, to the claims they make about their benchmark's ability to accurately measure the phenomenon. \Cref{fig:results_sankey} shows the results of key questions from the reviewing codebook which motivate our recommendations.

\paragraph{\textcolor{phenomenon}{Phenomenon}} 
The reviewed benchmarks cover a wide range of phenomena (\Cref{fig:results_overview} (A)), including areas such as reasoning (18.5\%), alignment (8.1\%) and code generation (5.7\%). 
% taxolomy root:
% NLP                      20.000000
% Reasoning                18.461538
% Agents                    8.791209
% Alignment                 8.131868
Most of the articles provided a definition of their measured phenomenon (78.2\%). Of the articles that provided definitions, 52.2\% of these definitions are widely agreed upon, but 47.8\% are contested, addressing phenomena with many possible definitions or no clear definition at all (\Cref{fig:results_sankey}).
For example, a benchmark measuring the extent to which LLM generations correspond to established psychometric categories  \cite{renValueBenchComprehensivelyEvaluating2024} has clear, widely agreed definitions. However, as a category, alignment benchmarks often target phenomena with contested definitions (e.g. `harmlessness').

The definitions of the phenomena also varied in whether they defined the phenomenon they tested as a composite (61.2\%) or a single unified whole  (36.5\%). For example, some phenomena can be tested alone, (e.g. measuring the ability to traverse a 2D map~\cite{nasirGameTraversalBenchmarkEvaluatingPlanning2024a}), while other phenomenon are overarching abilities integrating many sub-abilities (e.g., a model's `agentic capabilities' requiring sub-abilities such as intent recognition, alignment, and structured output generation~\cite{liuAgentBenchEvaluatingLLMs2024}).

% definition_integrity
% Composite phenomenon               61.233480
% Single cohesive phenomenon         36.563877
% Authors' description is unclear     2.202643

% what is the most common task -> task_ecology_clean
% percentage of tasks from benchmarks and/or exams task_source_clean
% Constructed	40.659341
% Representative	36.923077
% Partial	32.307692
% Complete	9.230769

% Constructed	40.659341
% Representative	36.923077
% Partial	32.307692
% Complete	9.230769
\paragraph{\textcolor{task}{Task}} 
The tasks chosen to measure the target phenomena varied widely, ranging from answering medical licensing exam questions~\cite{ouyangCliMedBenchLargescaleChinese2024a} and detecting errors in computer code~\cite{shahStackEvalBenchmarkingLlms2024a} to reconciling conflicting information on Wikipedia~\cite{houWikiContradictBenchmarkEvaluating2024a}. Less than 10\% of benchmarks used complete real-world tasks, such as writing a correct SQL query given a natural language query and a database structure~\cite{changDrspiderDiagnosticEvaluation2023}.
Overall, 40.7\% of all reviewed benchmarks make use of constructed tasks, such as reading fictional multi-party conversations and answering questions about the beliefs of the conversation participants to test `theory of mind'~\cite{kimFANToMBenchmarkStresstesting2023}, with 28.5\% using exclusively constructed tasks. 
Partially real-world tasks, such as accomplishing e-commerce tasks collected from real people on a mock website \cite{yaoWebShopScalableRealworld2022a}, and representative tasks, such as answering exam-style science questions \cite{wangSciBenchEvaluatingCollegelevel2024}, are used in 32.3\% and 36.9\% of reviewed benchmarks, respectively.

% task used at least partially:
% Constructed	40.659341
% Representative	36.923077
% Partial	32.307692
% Complete	9.230769

% 	task used exclusivley:	
% Constructed	28.571429	
% Representative	23.076923	
% Partial	21.758242	
% Complete	7.912088	

Benchmarks included task items from various sources, with only 33.6\% relying on a single source. Authors most commonly handcrafted new task items (43.3\%), followed by reusing data from existing benchmarks (42.6\%) and generating data with LLMs (31.2\%). Human exams and other pre-existing sources were used in 38.2\% of benchmarks. Among those with a single task source, the most common was another benchmark (7.7\% of benchmarks sourced their tasks solely from another benchmark).

% using as part of process:
% Author-crafted	43.296703
% Another benchmark	42.637363
% LLM-generated	31.208791
% Real task	29.670330
% Procedurally-generated	26.813187
% Crowd-sourced	19.120879
% Expert-crafted	12.967033
% Human exams	9.450549
% Unknown	0.879121

% only using:
% Another benchmark         7.692308
% Real task                 6.813187
% Author-crafted            6.593407
% Procedurally-generated    3.076923
% Expert-crafted            2.857143
% LLM-generated             2.417582

% Criterion sampling: creators take items from a larger set based on specified rules
% example: filtering recent ML papers to produce literature review questions (ajithLitSearchRetrievalBenchmark2024)
% Targeted sampling: creators define a task space and choose tasks within it strategically
% example: compiling news sources with differing opinions and asking the LLM to synthesise the information (huangEmbraceDivergenceRicher2024)
% Random sampling: creators define a task space and randomly sample from it
% example: collect samples of informal text from the Semantic Scholar Corpus and evaluate rewriting in academic prose (diaoDoolittleBenchmarksCorpora2023)

Task items within any benchmark are effectively a sample from a much larger conceptual set of possible items that could be used to operationalise the phenomenon. The methodology used to select this sample significantly impacts the benchmark's validity. In 12.3\% of cases, authors exclusively used readily accessible datasets as the source of their task items, a practice known as \emph{convenience} sampling \cite{alvi2016manual}. Another 27.0\% incorporated convenience sampling as part of their sampling strategy. Overall, 55.2\% of benchmarks used at least partially \emph{targeted} sampling, which involved defining a task space and strategically selecting items within it (e.g., compiling news sources with differing opinions and testing how well models can synthesise the information~\cite{huangEmbraceDivergenceRicher2024}). 46.2\% used \emph{criterion} sampling, where items are selected from a larger set based on specified rules, as part of their strategy (e.g., filtering recent machine learning articles to produce literature review questions~\cite{ajithLitSearchRetrievalBenchmark2024}). Finally, 17.1\% used \emph{random} sampling, which involves defining a task space and randomly selecting items from it (e.g., collecting samples of informal text from Semantic Scholar and evaluate rewriting in academic prose~\cite{diaoDoolittleBenchmarksCorpora2023a}). Of all reviewed articles, 17.5\% exclusively used targeted sampling, 10.8\% exclusively used criterion sampling, and 7.0\% exclusively used random sampling. 

% TO ADD: GIVE MORE INFO ON OTHER SAMPLING TECHNIQUES AND WHAT THEY MEAN, BASED ON THE DATA BELOW:

% partially using sampling method:
% Targeted	55.164835
% Criterion	46.153846
% Convenience	39.340659
% Random	17.142857
% Unknown	4.395604

% only using sampling method:
% Targeted       17.582418
% Convenience    12.307692
% Criterion      10.769231
% Random          7.032967

%{\color{red} TO DO: SIZE OF BENCHMARKS INFO}

% 4 most common response formats: response_format_clean: “structured response” is useful for the auxiliary tasks
	
% Free response	46.373626
% Multiple choice	40.000000
% Short free response	38.461538
% Structured	21.098901
% Interaction	6.593407
% Logits	1.318681

% what metric was featured how much: metric_definition_clean
% Exact match	81.318681
% Soft match	20.879121
% LLM-as-a-Judge	17.142857
% Human ratings	12.967033
% Distribution	12.087912
% LLM post-processing	9.450549
% Reward	8.791209
% Correlation	5.054945

Benchmarks included task items that require responses in a range of formats. Free response was the most common response format (used partially by 46.4\%, exclusively by 17.8\%), followed by multiple choice (used partially by 40.0\%, exclusively by 18.5\%). Short free response, where responses were limited to a few words, was used partially by 38.5\% and exclusively by 13.2\%, and other structured response formats such as JSON  were used partially by 21.1\% and exclusively by 8.6\%.

% response format used at least aprtially:
% Free response	46.373626
% Multiple choice	40.000000
% Short free response	38.461538
% Structured	21.098901
% Interaction	6.593407
% Logits	1.318681

% response format used exclusivley:
% Multiple choice        18.461538
% Free response          17.802198
% Short free response    13.186813
% Structured              8.571429
% Interaction             2.197802
% Logits                  1.098901

\paragraph{\textcolor{metric}{Metric}} 
The most common metric used to score the benchmarking tasks was exact matching (used at least partially by 81.3\%, exclusively by 40.7\%). Other commonly used metrics include soft match scores, which have an exact correct answer but allow for partial credit (used at least partially by 20.9\%, exclusively by 0.9\%), LLM-as-a-judge (at least partially by 17.1\%, exclusively by 3.1\%), and human ratings (at least partially by 13.0\%, exclusively by 1.8\%). Once the responses were scored, 16.0\% used uncertainty estimates or statistical tests to compare the results.

% percentage at least partially using metric:
% Exact match	81.318681
% Soft match	20.879121
% LLM-as-a-Judge	17.142857
% Human ratings	12.967033
% Distribution	12.087912
% LLM post-processing	9.450549
% Reward	8.791209
% Correlation	5.054945

% percentage only using metric
% Exact match            40.659341
% Reward                  3.296703
% LLM-as-a-Judge          3.076923
% Distribution            2.857143
% Human ratings           1.758242
% Soft match              0.879121
% Correlation             0.439560
% LLM post-processing     0.219780

% percent that compare results: results_comparison
% No     64.757709
% Yes    35.242291
\paragraph{\textcolor{claims}{Claims}}

To support their results, 53.4\% of articles presented evidence for the construct validity of their benchmark. This included benchmarks using real-world tasks to ensure the problems reflected actual coding~\cite{huangDAcodeAgentData2024} or authors suggesting that their designs better captured user intents~\cite{wangUsercentricMultiintentBenchmark2024}. 35.2\% of articles also included a comparison to other benchmarks of similar phenomena, 32.4\% to a human baseline, 31.2\% to a more realistic setting, enabling an understanding of similarities and differences.
\section{Recommendations}
Based on our review, we provide recommendations to strengthen the construct validity of LLM benchmarks. We prioritise broadly applicable recommendations, noting that not all apply to every benchmark. Each includes a specific checklist of items to address when creating a new benchmark.

\subsection{Define the phenomenon}
%Standard recommendation structure:

\begin{checklistbox}{Define the phenomenon}
    \begin{itemize}[leftmargin=5pt, label=\scriptsize$\square$]
        %\item The phenomenon being measured is clearly defined.
        \item Provide a precise and operational definition for the phenomenon being measured
        %\item The benchmark captures the entire targeted phenomenon, not a single aspect of it.
        \item Specify the scope of the phenomenon being covered and acknowledge any excluded aspects
        %\item When the benchmark aggregates sub-aspects of the phenomenon, they are described and measured separately.
        \item Identify if the phenomenon has sub-components and ensure they are measured separately
    \end{itemize}
\end{checklistbox}

%What is the desirable feature of benchmarks
The target phenomenon should be clearly defined before operationalising it in a benchmark. 78.2\% of reviewed benchmarks provide definitions, helping users to understand what is being measured and how. \textcite{phuongEvaluatingFrontierModels2024a} shows an example of how a clear definition can guide task design and clarify the operationalisation of an abstract concept.

% What can be done to improve?
When multiple definitions exist for a single term, stating one helps clarify the intention of the benchmark and how the results should be interpreted. If there is no consensus for defining a phenomenon, as we observe in 47.8\% of benchmarks, providing a negative or apophatic definition can help set boundaries (e.g., repeating memorised answers is not reasoning)~\cite{beanLINGOLYBenchmarkOlympiadlevel2024a}. If the target phenomenon has sub-components, as in 61.2\% of benchmarks, benchmarking each element separately increases clarity and improves interpretation. Although, benchmarks which combine several different measures of the same concept can be useful in no single measure is adequate.

\subsection{Measure the phenomenon and only the phenomenon}

\begin{checklistbox}{Measure only the phenomenon}
    \begin{itemize}[leftmargin=5pt, label=\scriptsize$\square$]
%\item Auxiliary tasks are either tested independently or their impact on the primary task is accounted for.
%\item Isolate or control for the impact of auxiliary tasks not central to the primary task
\item Control for unrelated tasks that may affect the results
%\item If the benchmark has a strict output format, model performance is compared with and without format constraints.
\item Assess the impact of format constraints on model performance
%\item If the benchmark uses automated techniques to parse answers, they are tested for bias and accuracy across different models.
\item Validate any automated output parsing techniques for accuracy, consistency and bias
\end{itemize}
\end{checklistbox}

%What is the desirable feature of benchmarks
Completing a benchmark task involves a combination of task-specific and more general abilities, such as instruction-following. Additional, unmeasured, subtasks can confound the measurement of the target phenomenon. For example, 21.1\% of benchmarks require specific output formats that can themselves be challenging for models~\cite{tamLetMeSpeak2024a}. Others may involve complex instructions that disproportionately reduce performance in weaker models~\cite{beanLINGOLYBenchmarkOlympiadlevel2024a}. In tasks such as commonsense reasoning, it can be difficult to separate reasoning ability from model's existing knowledge \cite{hoWikiWhyAnsweringExplaining2023}.

% What can be done to improve?
Several strategies can mitigate these confounding effects. Baselines can be established for performance on the relevant subtasks alone. If a benchmark requires world knowledge but does not intend to measure it, models should first be tested on this world knowledge directly and scores adjusted to avoid penalising failures arising from lack of knowledge. If a benchmark uses strict formats or complex instructions, test those skills independently and allow retries to distinguish formatting proficiency from task performance. If LLMs are used to parse original model responses, the extractor LLM should be validated to avoid introducing new biases or performance artifacts. Though less applicable on individual benchmarks, factor analysis techniques are also being explored to extract latent capability dimensions with less interference from auxiliary tasks \cite{ruan2024observational, ilic2024evidence, burnell2023revealing}.

\subsection{Construct a representative dataset for the task}
%Standard recommendation structure:
\begin{checklistbox}{Construct a representative dataset for the task}
    \begin{itemize}[leftmargin=5pt, label=\scriptsize$\square$]
 %\item Task items are chosen to be representative of the overall task space.
 \item Employ sampling strategies to ensure task items are representative of the overall task space
 %\item Checks are performed to ensure that the task items are high quality and related to the phenomenon (especially if the benchmark is large).
 \item Verify the quality and relevance of all task items especially for large or automatically generated datasets
 %\item The selection of task items has been tailored for testing LLMs (e.g. including both easy and difficult tasks by human standards, including small syntactic perturbations of task items).
 \item Include task items that test known LLM sensitivities (e.g. input permutations or variations)
\end{itemize}
\end{checklistbox}

%What is the desirable feature of benchmarks
Benchmarks use finite sets of task items as proxies for complex phenomena. Each item can be seen as drawn from a larger possible set, so sampling should be representative of the task space. However, 27.0\% of reviewed benchmarks used convenience sampling, relying on the validity of the existing sample. For example, if a benchmark reuses questions from a calculator-free exam such as AIME~\cite{whiteLiveBenchChallengingContaminationLimited2025}, numbers in each problem will have been chosen to facilitate basic arithmetic. Testing only on these problems would not predict performance on larger numbers, where LLMs struggle.

% What can be done to improve?
We recommend that authors adopt more robust sampling techniques, such as random or stratified sampling (17.1\% of reviewed benchmarks use at least one of these). With better sampling methods, smaller well-designed datasets can provide higher construct validity than larger datasets at less computational cost~\cite{maiapoloTinyBenchmarksEvaluatingLLMs2024}. The risk of having non-representative sampling of benchmark tasks should also be taken into account when generating synthetic examples to increase the size of benchmarks, as occurs in 47.5\% of the benchmarks we reviewed. Task items can also be explicitly designed to test for common weaknesses in LLMs. For example, human examinations are unlikely to have the same question repeated in several different phrasings, but LLMs are known to be sensitive to minor variations in prompts \cite{zhuoProSAAssessingUnderstanding2024, mizrahiStateWhatArt2024} and variations could improve the robustness of the results.

\subsection{Acknowledge limitations of reusing datasets}
%Standard recommendation structure:
\begin{checklistbox}{Acknowledge limitations of reusing datasets}
    \begin{itemize}[leftmargin=5pt, label=\scriptsize$\square$]
    %\item The benchmark documents whether it adapts a previous dataset or benchmark.
    \item Document whether the benchmark adapts a previous dataset or benchmark
    %\item If so, the authors provide a clear analysis of the previous work describing strengths and limitations of the benchmark.
    \item If so, analyse and report the relevant strengths and limitations of the adapted prior work
    %\item If so, results on the original benchmark are included and compared to.
    \item If so, report and compare performance on the new benchmark against the original
    %\item Any differences from the original dataset are justified and explained in the context of construct validity.
    %\item Articulate the rationale when modifying a reused dataset, identifying improvements to construct validity
    \item Explain modifications to reused datasets and how they improve construct validity
\end{itemize}
\end{checklistbox}

%What is the desirable feature of benchmarks
38.2\% of reviewed benchmarks reuse data from previous benchmarks or human exams. Reusing existing datasets makes it difficult for authors to control the construct validity of their benchmark and limits options for design choices such as task and metric. Reuse of existing materials also increases the chances of benchmarking tasks appearing in pre-training data (see \Cref{rec:contamination}), compromising results.

% What can be done to improve?

Newly constructed datasets should be preferred to reused datasets.
When datasets are reused, such as when a benchmark improves upon an older version (7.7\% of cases), authors must investigate which changes have been introduced and what the new benchmark preserves.
Differences between the original and new datasets should be clearly documented and justified. Reporting differences in results between the new and original benchmarks can help to demonstrate the impact of changes, including whether the new benchmark has improved the construct validity.

\subsection{Prepare for contamination}
\label{rec:contamination}
\begin{checklistbox}{Prepare for contamination}
    \begin{itemize}[leftmargin=5pt, label=\scriptsize$\square$]
    %\item The benchmark includes a method for identifying contamination.
    \item Implement tests to detect data contamination and apply them to the benchmark
    %\item Held-out task items are available (e.g. via a private test set, or generating new questions).
    \item Maintain a held-out set of task items to facilitate ongoing, uncontaminated evaluation
    %\item The authors conduct an analysis of data exposure prior to the benchmark creation.
    \item Investigate the potential pre-exposure of benchmark source materials or similar data in common LLM training corpora
\end{itemize}
\end{checklistbox}

For many phenomena, the process through which an LLM reaches the answer is equally as important as whether the correct answer was reached. In these cases, the validity of the results can be undermined both by direct contamination of benchmark items and by memorisation of partial answers or closely-related information. Benchmark contamination is likely to occur even with model developers acting in good faith \cite{zhouDontMakeYour2023}. 
When a benchmark is widely used, the progressive effort to improve on that benchmark means that newly developed techniques will be specifically suited to solving that task. Over time, this selection of methods can lead to overfitting, similar to the repeated use of a validation set \cite{zhangCarefulExaminationLarge2024a}, effectively contaminating the benchmark.

We recommend vetting test items for dataset contamination when the benchmark is created, especially when the dataset is already public, or when an LLM is used to generate task examples \cite{mainiLLMDatasetInference2024}. Including these contamination checks within the benchmark itself can provide ongoing verification of the validity. Dynamic benchmarks have also been proposed as a solution to this issue \cite{kielaDynabenchRethinkingBenchmarking2021d}, and procedurally generated tasks, in particular, can be used to keep benchmarks up-to-date \cite{khoujaLINGOLYTOODisentanglingMemorisation2025}.

\subsection{Use statistical methods to compare models}
\begin{checklistbox}{Use statistical methods to compare models}
    \begin{itemize}[leftmargin=5pt, label=\scriptsize$\square$]
     %\item The sample size of the benchmark is reported and sufficiently powered.
     \item Report the benchmark's sample size and justify its statistical power
     %\item Uncertainty estimates are provided for the main scores, and are narrow enough to meaningfully compare relevant models.
     \item Report uncertainty estimates for all primary scores to enable robust model comparisons
     %\item If human raters are used, the recruitment accounts for demographic biases which may be relevant to the preferences they report.
     \item If using human raters, describe their demographics and mitigate potential demographic biases in rater recruitment and instructions
    %\item If the benchmark uses subjective measures of performance, the distribution of labels is reported and accounted for in the scoring.
    \item Use metrics that capture the inherent variability of any subjective labels, without relying on single-point aggregation or exact matching

\end{itemize}
\end{checklistbox}

To support valid interpretation and comparisons across models, prior work has highlighted the importance of using statistical techniques in the analysis of benchmark results~\cite{millerAddingErrorBars2024,mcintoshInadequaciesLargeLanguage2024}. At present, only 16.0\% of reviewed benchmarks conducted any statistical testing. Increasing the use of robust statistical methods for LLM benchmarking is critical.

In addition, scoring methods based on human or LLM ratings provide subjective metrics that may vary across samples. Since there is real variation in human preferences, the aggregation and reporting should consider the meaning of the distribution of ratings. In particular, benchmark creators should consider the representativeness of the raters, and whether there are meaningful differences between groups \cite{kirkPRISMAlignmentDataset2024}. For example, bias benchmarks operate with concepts of harm and bias which are culturally and socially contingent \cite{sahooIndiBiasBenchmarkDataset2024}. By considering the distribution of ratings, overall results will better incorporate potential real-world uses of LLMs.

\subsection{Conduct an error analysis}
\begin{checklistbox}{Conduct an error analysis}
    \begin{itemize}[leftmargin=5pt, label=\scriptsize$\square$]
    %\item The authors include an analysis of failure cases.
    \item Conduct a qualitative and quantitative analysis of common failure modes
    %\item There are no patterns of failures which relate to non-targeted phenomena.
    \item Investigate whether failure modes correlate with non-targeted phenomena (confounders) rather than the intended construct
    %\item If so, potential biases in the scoring are discussed.
    \item If so, identify and discuss any potential scoring biases revealed in the error analysis
    %\item Experiments or recommendations are made to improve model scores on the benchmark.
    %\item Conduct experiments or propose new directions to improve model scores on the benchmark
\end{itemize}
\end{checklistbox}

After a benchmark is created, an error analysis can reveal the types of errors models make. If the benchmark has high construct validity, these errors will indicate useful research directions for the target phenomenon. Therefore, error analysis can provide an indication of the construct validity of the benchmark based on the avenues for improvement which are indicated. If the failure cases correspond to failures to demonstrate the target phenomenon, the validity is high. If not, this may be a reason to modify the benchmark to be more precise. As an example, \citeauthor{phuongEvaluatingFrontierModels2024a} experiment with repeated trials and find that the highest scores come from low probability generations, allowing them to identify that the tasks are possible for the models, but not likely to be solved.

\subsection{Justify construct validity}
\vspace{-5pt}
\begin{checklistbox}{Justify construct validity}
    \begin{itemize}[leftmargin=5pt, label=\scriptsize$\square$]
    %\item A relevant use case is provided to justify the direction of research.
    \item Justify the relevance of the benchmark for the phenomenon with real-world applications
    %\item The authors explain why the task and metric were chosen to measure the target phenomenon.
    \item Provide a clear rationale for the choice of tasks and metrics, connected to the operational definition of the phenomenon
    %\item The benchmark is compared to other benchmarks of similar phenomena, with a discussion of similarities and differences.
    \item Compare similarities and differences between the benchmark and existing evaluations of similar phenomena
    \item Discuss the limitations and design trade-offs of the benchmark concerning construct validity
\end{itemize}
\end{checklistbox}

Improving scores on a benchmark requires attention and resources, so it is helpful for the authors to explain why the benchmark is relevant. However, only about half (53.4\%) of reviewed benchmarks justify why they are a valid measure of an important phenomenon. To establish high construct validity, we recommend authors articulate the rationale behind the chain of decisions from defining a phenomenon, to operationalising it via a task, to selecting specific task items to test, to the code implementation of the task, up to making validity claims. 

Authors should discuss key design trade-offs. For example, multiple-choice formats are easy to score but can be gamed and rarely reflect real-world use cases~\cite{mccoyRightWrongReasons2019b}. Free-text responses are more realistic but harder and costlier to evaluate~\cite{bidermanLessonsTrenchesReproducible2024a}. With many benchmarks aiming at similar phenomena (e.g. `reasoning'), clarity about how a benchmark aligns or diverges from others is critical. Convergent and discriminant validity help clarify what each benchmark actually tests~\cite{alaaMedicalLargeLanguage2025a}. These choices should be addressed directly in the limitations to enable more reliable and interpretable progress.

\begin{examplebox}{Example}
As a practical demonstration of our recommendations, we discuss the GSM8K benchmark \cite{cobbe_training_2021}. We chose this benchmark because of its widespread adoption and examples of simple ways to address many, but not all, of our recommendations.\\

\textbf{Define the phenomenon:} 
GSM8K describes itself as `grade school math problems...using basic arithmetic' which are `useful for probing the informal reasoning ability
 of large language models'. This definition describes both the phenomenon, [mathematical] reasoning, and the task, arithmetic-based math problems, making the operationalisation of `reasoning' and the scope of the assessment clear. The authors also note that the difficulty comes from a combination of `reading comprehension' and `logical reasoning'. Based on our checklist, we would recommend assessing the relative impact of each of these skills. One approach would be to annotate the relative demand for each skill on each of the problems, similar to \citeauthor{zhou2025general} \cite{zhou2025general}.\\

\textbf{Measure only the phenomenon:} 
Beyond reading comprehension and logical reasoning, GSM8K requires arithmetic calculations. The authors note this as a potential confounder and provide a calculator tool to their model. However, we would recommend building this into the benchmark directly to avoid score differences due to testing setups \cite{bidermanLessonsTrenchesReproducible2024a}. Instead, the answer could be provided as a mathematical expression which is evaluated as part of the scoring. The answer format itself is simple, although the potential impacts of tokenization should be considered \cite{singh2024tokenization}.\\

\textbf{Construct a representative dataset for the task:} 
Annotators creating the dataset are given a diverse set of seed prompts and the dataset is tested for reliability by human annotators prior to release. Additional tests to target known LLM weaknesses, such as re-wording the questions, would improve the robustness of the results~\cite{mirzadehGSMSymbolicUnderstandingLimitations2024}.\\

\textbf{Acknowledge limitations of reusing datasets:} 
The dataset is created from scratch for this project, which avoids the issues of reused datasets.\\

\textbf{Prepare for contamination:} 
We recommend the addition of a canary string and a set of held-out task items. Testing performance on a new set of similar benchmark questions also shows significant performance drops, likely indicating that contamination has occurred \cite{zhang2024careful}.\\

\textbf{Use statistical methods to compare models:} 
The benchmark reports a large sample size justified by the deliberate creation of diverse questions. Comparisons between the models being tested are reported over multiple runs with standard deviations.\\

\textbf{Conduct an error analysis:} 
No error analysis is conducted. By creating a typology of error types on GSM8K, other works were able to guide improvements in future models \cite{wei2022chain, wang-etal-2023-plan}.\\

\textbf{Justify construct validity:} 
The authors describe how their dataset differs from existing sets of maths problems in quality and scale. We would also recommend a discussion of how maths reasoning problems relate to logical reasoning broadly given the stated requirements of `reading comprehension' and `logical reasoning'.\\

\textbf{Conclusion:}
GSM8K is a generally valid benchmark for measuring performance on grade school maths questions. Contamination has likely been a factor in increased scores over time which may have been partially avoidable. An error analysis would also have furthered the usefulness of the benchmark. Where interpretation stretches from `math problems' to `logical reasoning', greater clarity would be valuable to help readers interpret the results.

\end{examplebox}
\section{Discussion}
\label{sec:limitations}
%This needs to go somewhere for the checklist.
%\subsection{Principle findings}

We performed a systematic review of 445 benchmarks from the NLP and ML literature to assess the best practices around construct validity. We found that the operationalisation of abstract phenomena was often insufficient, with definitions being missing or contested. Tasks were frequently taken from pre-existing data sources without adjustments to ensure that they were representative of the target phenomenon. Statistical testing was also rarely performed. About half of the reviewed articles did discuss the validity of their benchmark, but nearly every paper had weaknesses in at least one area. In light of these gaps, we created a list of recommendations covering the design of phenomena, tasks, and metrics as well as interpretation to improve the construct validity of future LLM benchmarks.

%\subsection{How to use the Checklist}
We developed the operational checklist as a practical tool to support researchers in proactively engaging with construct validity throughout the benchmark lifecycle. We recommend its use early in the design phase to guide critical design decisions regarding task selection, sample construction, metric justification, as well as later when interpreting findings. We do not expect that every benchmark will satisfy every item, as practical trade-offs will sometimes be necessary. We recommend reporting the checklist as an appendix with answers and explanations for each skipped item, allowing users to assess if a benchmark aligns with their needs and matches best practices.
Beyond new benchmark developments, the checklist can also serve as an evaluation framework for existing benchmarks, or for adapting them to new domains or capabilities.

%\subsection{Limitations}\label{sec:limitations}

% \paragraph{Selection Bias} 
We describe limitations of our approach. Our focus on leading conference proceedings, while ensuring a baseline of peer-reviewed quality, may systematically exclude certain types of impactful benchmarks. For example, it does not capture benchmarks developed and released by industry labs without formal peer review, or those published in specialised domain-specific venues, may not be captured. We primarily review benchmarks prevalent in mainstream academic AI research.

% \paragraph{LLM Filtering} 
To manage the extensive initial corpus, we employed GPT-4o mini for preliminary screening against topic, empirical, and modality criteria, prior to manual review. This automated step was only used to exclude articles, and validated against human annotation (see~\cref{app:schema} indicating good agreement). Nevertheless, this may have introduced undetected false negative systematic errors. Distributional shift in language usage may also have contributed to the lower inclusion of older papers, alongside the general increase in papers being published in this area. The scale of the review also necessitated limiting the number of reviewers per paper, reducing the robustness of the reviews.

\section{Conclusion}
The rapid advancement of LLMs requires robust evaluation. Our systematic review of 445 benchmarks reveals prevalent gaps that undermine the construct validity needed to accurately measure targeted phenomena. To address these shortcomings, which can hinder genuine progress, we propose eight recommendations and a practical checklist for designing and interpreting LLM benchmarks. Ultimately, ``measuring what matters'' requires a conscious, sustained effort from the research community to prioritise construct validity, fostering a cultural shift towards more explicit and rigorous validation of evaluation methodologies.

%%%%%%%%  CLOSING MATTER %%%%%%%%%%%%%%%%
\begin{ack}
A.M.B. is supported in part by the Clarendon Scholarship and the Dieter Schwarz foundation. R.O.K. is supported by the Clarendon Scholarship, the Jesus College Old Members' Scholarship, and the Cosmos Fellowship. H.M. is supported by ESRC [ES/P000649/1] and would like to acknowledge the London Initiative for Safe AI. C.E. is supported by the EPSRC Centre for Doctoral Training in Health Data Science (EP/S02428X/1) and the AXA Research Fund. F.L. is supported by Clarendon and Jason Hu studentships. H.R.K.’s PhD is supported by the Economic and Social Research Council grant ES/P000649/1. M.G. is supported by the SMARTY (PCI2024-153434) project funded by the Agencia Estatal de Investigación (doi:10.13039/501100011033) and by the European Commission through the Chips Act Joint Undertaking project SMARTY (Grant 101140087). This material is based in part upon work supported by the National Science Foundation Graduate Research Fellowship Program under Grant No. DGE-2139841. O.D. is supported by the UKRI’s EPSRC AIMS CDT grant (EP/S024050/1). J.R's PhD is supported by the Engineering and Physical Sciences Research Council [Grant Number EP/W524311/1]. J.B. acknowledges the German Federal Ministry of Research, Technology, and Space (16DII131). A. Bibi would like to acknowledge the UK AISI systemic safety grant. A. Bosselut gratefully acknowledges the support of the Swiss National Science Foundation (No. 215390), Innosuisse (PFFS-21-29), the EPFL Center for Imaging, Sony Group Corporation, and a Meta LLM Evaluation Research Grant. This work is supported by the UKRI grant: Turing AI Fellowship EP/W002981/1. L.R. acknowledges support from the Royal Society Research Grant RG{\textbackslash}R2{\textbackslash}232035 and the UKRI Future Leaders Fellowship [MR/Y015711/1]. Any opinions, findings, and conclusions or recommendations expressed in this material are those of the author(s) and do not necessarily reflect the views of the National Science Foundation.
\end{ack}

\clearpage
\printbibliography
\end{refsection} 

%% CHECK LIST %%

%%%%%%%%  SUPPLEMENTARY MATERIAL %%%%%%%%%%%%%%%%
\appendix
\begin{refsection}

% Comment this out for ArXiv
%
%\input{sections/neurips_checklist}

\clearpage
\part{Supplementary Material} % Start the appendix part
\mtcsetdepth{parttoc}{1}
\parttoc % Insert the appendix TOC

%\addcontentsline{toc}{section}{\large{PART I: Dataset Details and Distributions}}

\clearpage
\section{Construct Validity Checklist}

For usability, we have reproduced the complete construct validity checklist here, grouped by recommendation. We recommend that checklist users consider each of these questions and whether the answer is adequately addressed in the benchmark and corresponding paper. We anticipate that it may be difficult to adopt every recommendation in every case. For example, computing confidence intervals may be prohibitively expensive. In these cases, considering and discussing the tradeoffs of adopting the recommendations as limitations would enable readers of these papers to better interpret the results.

\textbf{Define the phenomenon}
\begin{itemize}[leftmargin=25pt, label=\scriptsize$\square$]
        %\item The phenomenon being measured is clearly defined.
        \item Provide a precise and operational definition for the phenomenon being measured
        %\item The benchmark captures the entire targeted phenomenon, not a single aspect of it.
        \item Specify the scope of the phenomenon being covered and acknowledge any excluded aspects
        %\item When the benchmark aggregates sub-aspects of the phenomenon, they are described and measured separately.
        \item Identify if the phenomenon has sub-components and ensure they are measured separately
    \end{itemize}

\textbf{Measure only the phenomenon}
\begin{itemize}[leftmargin=25pt, label=\scriptsize$\square$]
    \item Control for unrelated tasks that may affect the results
%\item If the benchmark has a strict output format, model performance is compared with and without format constraints.
\item Assess the impact of format constraints on model performance
%\item If the benchmark uses automated techniques to parse answers, they are tested for bias and accuracy across different models.
\item Validate any automated output parsing techniques for accuracy, consistency and bias

\end{itemize}

\textbf{Construct a representative dataset for the task}
\begin{itemize}[leftmargin=25pt, label=\scriptsize$\square$]
    \item Employ sampling strategies to ensure task items are representative of the overall task space
 %\item Checks are performed to ensure that the task items are high quality and related to the phenomenon (especially if the benchmark is large).
 \item Verify the quality and relevance of all task items, especially for large or automatically generated datasets
 %\item The selection of task items has been tailored for testing LLMs (e.g. including both easy and difficult tasks by human standards, including small syntactic perturbations of task items).
 \item Include task items that test known LLM sensitivities (e.g. input permutations or variations)
\end{itemize}

\textbf{Acknowledge limitations of reusing datasets}
\begin{itemize}[leftmargin=25pt, label=\scriptsize$\square$]
    \item Document whether the benchmark adapts a previous dataset or benchmark
    %\item If so, the authors provide a clear analysis of the previous work describing strengths and limitations of the benchmark.
    \item If so, analyse and report the relevant strengths and limitations of the adapted prior work
    %\item If so, results on the original benchmark are included and compared to.
    \item If so, report and compare performance on the new benchmark against the original
    %\item Any differences from the original dataset are justified and explained in the context of construct validity.
    %\item Articulate the rationale when modifying a reused dataset, identifying improvements to construct validity
    \item Explain modifications to reused datasets and how they improve construct validity
\end{itemize}

\textbf{Prepare for contamination}
\begin{itemize}[leftmargin=25pt, label=\scriptsize$\square$]
    \item Implement tests to detect data contamination and apply them to the benchmark
    %\item Held-out task items are available (e.g. via a private test set, or generating new questions).
    \item Maintain a held-out set of task items to facilitate ongoing, uncontaminated evaluation
    %\item The authors conduct an analysis of data exposure prior to the benchmark creation.
    \item Investigate the potential pre-exposure of benchmark source materials or similar data in common LLM training corpora
\end{itemize}

\textbf{Use statistical methods to compare models}
\begin{itemize}[leftmargin=25pt, label=\scriptsize$\square$]
     \item Report the benchmark's sample size and justify its statistical power
     %\item Uncertainty estimates are provided for the main scores, and are narrow enough to meaningfully compare relevant models.
     \item Report uncertainty estimates for all primary scores to enable robust model comparisons
     %\item If human raters are used, the recruitment accounts for demographic biases which may be relevant to the preferences they report.
     \item If using human raters, describe their demographics and mitigate potential demographic biases in rater recruitment and instructions
    %\item If the benchmark uses subjective measures of performance, the distribution of labels is reported and accounted for in the scoring.
    \item Use metrics that capture the inherent variability of any subjective labels, without relying on single-point aggregation or exact matching.
\end{itemize}

\textbf{Conduct an error analysis}
\begin{itemize}[leftmargin=25pt, label=\scriptsize$\square$]
    \item Conduct a qualitative and quantitative analysis of common failure modes
    %\item There are no patterns of failures which relate to non-targeted phenomena.
    \item Investigate whether failure modes correlate with non-targeted phenomena (confounders) rather than the intended construct
    %\item If so, potential biases in the scoring are discussed.
    \item If so, identify and discuss any potential scoring biases revealed in the error analysis
    %\item Experiments or recommendations are made to improve model scores on the benchmark.
    \item Conduct experiments or propose new directions to improve model scores on the benchmark
\end{itemize}

\textbf{Justify construct validity}
\begin{itemize}[leftmargin=25pt, label=\scriptsize$\square$]
    \item Justify the relevance of the benchmark for the phenomenon with real-world applications
    %\item The authors explain why the task and metric were chosen to measure the target phenomenon.
    \item Provide a clear rationale for the choice of tasks and metrics, connected to the operational definition of the phenomenon
    %\item The benchmark is compared to other benchmarks of similar phenomena, with a discussion of similarities and differences.
    \item Compare similarities and differences between the benchmark and existing evaluations of similar phenomena
    \item Discuss the limitations and design trade-offs of the benchmark concerning construct validity
\end{itemize}

\clearpage
\section{Complete Codebook}
\label{app:codebook}

This section describes each of the items in the codebook with summaries of the results where possible. The complete codebook is available as a dataset on \href{https://huggingface.co/datasets/ambean/construct-validity-review}{Hugging Face} and the code used to clean the dataset is available on \href{https://github.com/am-bean/benchmark_review}{GitHub}.

\subsection{General Background and Summary}
{\small\begin{longtable}{p{.005\textwidth}p{.995\textwidth}}
\multicolumn{2}{l}{\textbf{bibkey}} \\
& \textit{Description}: The unique identifier to match the reviewed paper to a \texttt{.bib} file. \\
& \textit{Codebook question}: ID of article (this is provided in the list of papers) \\
\\
\multicolumn{2}{l}{\textbf{title}} \\
& \textit{Description}: The title of the article. \\
& \textit{Codebook question}: Title of article \\
\\
\multicolumn{2}{l}{\textbf{benchmark}} \\
& \textit{Description}: The name of the benchmark. \\
& \textit{Codebook question}: The name of the benchmark, if one exists (e.g. GSM8K) \\
\\
\multicolumn{2}{l}{\textbf{inclusion}} \\
& \textit{Description}: Whether the paper was included in the review. \\
& \textit{Codebook question}: According to the criteria, should this paper be included or excluded? \\
\\
\multicolumn{2}{l}{\textbf{exclusion\_criteria}} \\
& \textit{Description}: The criteria for excluding the paper, if any. \\
& \textit{Codebook question}: If exclude, what criteria is violated? \\
\\
\multicolumn{2}{l}{\textbf{exclusion\_criteria\_detail}} \\
& \textit{Description}: Any additional details about the exclusion criteria. \\
& \textit{Codebook question}: If exclude, why? (optional, 1 sentence) \\
\\
\multicolumn{2}{l}{\textbf{short\_summary}} \\
& \textit{Description}: A short summary of the paper. \\
& \textit{Codebook question}: Short summary of paper contribution and method. Likely to be similar to the abstract. (2-3 sentences, no need for numbers) \\
\\
\multicolumn{2}{l}{\textbf{contribution}} \\
& \textit{Description}: Any additional notes about the article contribution \\
& \textit{Codebook question}: Other useful notes on contribution details (optional, only if something stood out) \\
\\
\end{longtable}}

\subsection{\textcolor{phenomenon}{Phenomenon}}
{\small\begin{longtable}{p{.005\textwidth}p{.995\textwidth}}
\multicolumn{2}{l}{\textbf{target\_phenomenon}} \\
& \textit{Description}: The main phenomenon measured in the paper, as defined by the authors. \\
& \textit{Codebook question}: According to the authors, what capability or specific application is being measured? (a few words, e.g. knowledge, reasoning, natural language understanding) \\
\\
\multicolumn{2}{l}{\textbf{phenomenon\_short}} \\
& \textit{Description}: Whether the phenomenon is a general capability or a specific application. \\
& \textit{Codebook question}: Which category does the target phenomenon fall into? \\
\\
& \textit{Summary of values:} \\
& General Capability (A broadly useful ability, which could be relevant to multiple applications): 321 \\
& Specific Application (A single use case, where the benchmark is likely to be examples of that use case): 118 \\
& Other: 16 \\
\\
\multicolumn{2}{l}{\textbf{phenomenon\_defined}} \\
& \textit{Description}: Whether the phenomenon is defined in the paper. \\
& \textit{Codebook question}: Is the targeted phenomenon explicitly defined? \\
\\
& \textit{Summary of values:} \\
& Yes: 348 \\
& No: 99 \\
\\
\multicolumn{2}{l}{\textbf{phenomenon\_definition}} \\
& \textit{Description}: The definition of the phenomenon. \\
& \textit{Codebook question}: How is the phenomenon of interest defined? (copy paste if possible, otherwise summarise what is being said) \\
\\
\multicolumn{2}{l}{\textbf{phenomenon\_taxonomy\_root}} \\
& \textit{Description}: The root category of the phenomenon taxonomy. \\
& \textit{Codebook question}: None \\
\\
\multicolumn{2}{l}{\textbf{phenomenon\_taxonomy\_leaf}} \\
& \textit{Description}: The leaf category of the phenomenon taxonomy. \\
& \textit{Codebook question}: None \\
\\
\multicolumn{2}{l}{\textbf{phenomenon\_taxonomy\_alternate}} \\
& \textit{Description}: An alternate for the phenomenon taxonomy if highly relevant. \\
& \textit{Codebook question}: None \\
\\
\multicolumn{2}{l}{\textbf{phenomenon\_contested}} \\
& \textit{Description}: Whether the definition of the phenomenon is broadly agreed upon, or if many definitions exist for the same term. \\
& \textit{Codebook question}: Does the target phenomenon have a widely agreed-upon definition, or is this definition contested? \\
\\
& \textit{Summary of values:} \\
& Contested: 225 \\
& Widely-agreed: 203 \\
& Not defined: 27 \\
\\
\multicolumn{2}{l}{\textbf{phenomenon\_contested\_clean}} \\
& \textit{Description}: Standardised mapping of phenomenon\_contested values for statistical analysis. \\
& \textit{Codebook question}: None \\
\\
& \textit{Summary of values:} \\
& ['Contested']: 225 \\
& ['Widely-agreed']: 203 \\
& ['No definition']: 27 \\
\\
\multicolumn{2}{l}{\textbf{definition\_scope}} \\
& \textit{Description}: Whether the benchmark covers everything within the phenomenon definition or only a subset. \\
& \textit{Codebook question}: Does the benchmark claim to measure everything covered by the definition, or focus on a more specific case or subset? \\
\\
& \textit{Summary of values:} \\
& Subset: 253 \\
& Comprehensive: 196 \\
\\
\multicolumn{2}{l}{\textbf{definition\_integrity}} \\
& \textit{Description}: Whether the definition is described as containing separate sub-phenomena. \\
& \textit{Codebook question}: Do the authors describe the phenomena as a single cohesive whole, or does it consist of sub-elements? \\
\\
& \textit{Summary of values:} \\
& Composite phenomenon: 278 \\
& Single cohesive phenomenon: 166 \\
& Other: 10 \\
\\
\multicolumn{2}{l}{\textbf{definition\_integrity\_detail}} \\
& \textit{Description}: If the definition includes sub-elements, what are they? \\
& \textit{Codebook question}: If the target phenomenon consists of sub-elements, are they measured separately? \\
\\
& \textit{Summary of values:} \\
& Yes: 261 \\
& Not applicable: 144 \\
& No: 46 \\
\\
\multicolumn{2}{l}{\textbf{purpose\_extra}} \\
& \textit{Description}: Any additional notes about the conceptual details of the paper. \\
& \textit{Codebook question}: Other useful notes on conceptual details (optional, only if something stood out) \\
\\
\end{longtable}}

\subsection{\textcolor{task}{Task} and Dataset}
{\small\begin{longtable}{p{.005\textwidth}p{.995\textwidth}}
\multicolumn{2}{l}{\textbf{task\_definition}} \\
& \textit{Description}: The definition of the benchmarking task. \\
& \textit{Codebook question}: How is the task defined? (1-2 sentences) \\
\\
\multicolumn{2}{l}{\textbf{task\_face\_validity}} \\
& \textit{Description}: An assessment of the face validity of the benchmark. \\
& \textit{Codebook question}: Is there prima facie reason to believe that this task could benchmark the target phenomenon? \\
\\
\multicolumn{2}{l}{\textbf{task\_face\_validity\_clean}} \\
& \textit{Description}: Standardised mapping of task\_face\_validity values for statistical analysis. \\
& \textit{Codebook question}: None \\
\\
\multicolumn{2}{l}{\textbf{task\_item\_definition}} \\
& \textit{Description}: The definition and/or an example of a single item in the task. \\
& \textit{Codebook question}: What does a single item in the task dataset look like? (If the task is stored as a table, what is represented by one row in the table?) (1-2 sentences) \\
\\
\multicolumn{2}{l}{\textbf{task\_definition\_detail}} \\
& \textit{Description}: Any additional notes about the task definition. \\
& \textit{Codebook question}: Any additional details on task definition. (optional, only if something stands out) \\
\\
\multicolumn{2}{l}{\textbf{task\_source}} \\
& \textit{Description}: The source of the task items. \\
& \textit{Codebook question}: What is the source of the dataset task items? (Choose all that apply. If additional comments are needed, use the next question.) \\
\\
\multicolumn{2}{l}{\textbf{task\_source\_clean}} \\
& \textit{Description}: Standardised mapping of task\_source values for statistical analysis. \\
& \textit{Codebook question}: None \\
\\
\multicolumn{2}{l}{\textbf{task\_source\_detail}} \\
& \textit{Description}: Any additional notes about the task source. \\
& \textit{Codebook question}: Other useful notes on task source (optional, use this is something needs to be clarified) \\
\\
\multicolumn{2}{l}{\textbf{task\_ecology}} \\
& \textit{Description}: How closely does the benchmarking task resemble the real application? \\
& \textit{Codebook question}: Is the task ecologically valid? (e.g. would a person really use a model in this way?) In the case of benchmarks which cover foundational abilities across many potential applications, you may need to select multiple responses and clarify below. \\
\\
\multicolumn{2}{l}{\textbf{task\_ecology\_clean}} \\
& \textit{Description}: Standardised mapping of task\_ecology values for statistical analysis. \\
& \textit{Codebook question}: None \\
\\
\multicolumn{2}{l}{\textbf{task\_ecology\_detail}} \\
& \textit{Description}: Any additional detail about the ecological validity of the task \\
& \textit{Codebook question}: Any additional detail about the ecological validity of the task \\
\\
\multicolumn{2}{l}{\textbf{task\_train\_val}} \\
& \textit{Description}: The data splits that are provided. \\
& \textit{Codebook question}: Which of the following dataset splits are provided? (if no splits are provided, assume the entire task is the test set) \\
\\
& \textit{Summary of values:} \\
& Test: 275 \\
& Test, Train, Validation: 96 \\
& Test, Train: 51 \\
& Test, Validation: 17 \\
& Other: 2 \\
\\
\multicolumn{2}{l}{\textbf{task\_dataset\_size}} \\
& \textit{Description}: The numbers of items in the task test dataset. \\
& \textit{Codebook question}: Size of the task dataset (count, test set only, if none is reported write "NA") \\
\\
\multicolumn{2}{l}{\textbf{task\_dataset\_size\_extra}} \\
& \textit{Description}: The number of items in the task train and validation datasets, if they exist. \\
& \textit{Codebook question}: The size of the train and validation splits, if they are provided \\
\\
\multicolumn{2}{l}{\textbf{task\_dataset\_size\_detail}} \\
& \textit{Description}: Any additional notes about the task dataset size. \\
& \textit{Codebook question}: Any additional notes (e.g. the test set for some of the subcategories is very small) \\
\\
\multicolumn{2}{l}{\textbf{task\_dataset\_metadata}} \\
& \textit{Description}: Whether additional metadata is provided about the task items. \\
& \textit{Codebook question}: Does the dataset provide any metadata? (e.g. topic area, difficulty level. Do not look in the dataset, this must be described in the paper to count.) \\
\\
& \textit{Summary of values:} \\
& Yes: 322 \\
& No: 126 \\
\\
\multicolumn{2}{l}{\textbf{dataset\_metadata\_detail}} \\
& \textit{Description}: A description of any metadata provided. \\
& \textit{Codebook question}: If metadata is provided, what is it? (comma-separated list of fields, e.g. human difficulty, date, language) \\
\\
\multicolumn{2}{l}{\textbf{dataset\_sampling\_method}} \\
& \textit{Description}: The method by which task items were selected from the space of possible task items. \\
& \textit{Codebook question}: How does the dataset relate to the population it represents? (Choose all that apply, see the image for examples) \\
\\
\multicolumn{2}{l}{\textbf{dataset\_sampling\_method\_clean}} \\
& \textit{Description}: Standardised mapping of dataset\_sampling\_method values for statistical analysis. \\
& \textit{Codebook question}: None \\
\\
\multicolumn{2}{l}{\textbf{response\_format}} \\
& \textit{Description}: The format of the expected response. \\
& \textit{Codebook question}: What is the format of the expected response? (Choose all that apply. Try to stick with the provided categories, and use the next question to clarify.) \\
\\
\multicolumn{2}{l}{\textbf{response\_format\_clean}} \\
& \textit{Description}: Standardised mapping of response\_format values for statistical analysis. \\
& \textit{Codebook question}: None \\
\\
\multicolumn{2}{l}{\textbf{response\_format\_detail}} \\
& \textit{Description}: Any additional notes about the response format. \\
& \textit{Codebook question}: Any additional details about the required response format to clarify how it fits in the categories above (optional) \\
\\
\end{longtable}}

\subsection{\textcolor{metric}{Metric}}
{\small\begin{longtable}{p{.005\textwidth}p{.995\textwidth}}
\multicolumn{2}{l}{\textbf{metric\_definition}} \\
& \textit{Description}: The definition of the metric used to score the benchmark. \\
& \textit{Codebook question}: What is the primary metric for scoring the benchmark? (Choose all that apply. Please try to stick to the provided categories and elaborate below.) \\
\\
\multicolumn{2}{l}{\textbf{metric\_definition\_clean}} \\
& \textit{Description}: Standardised mapping of metric\_definition values for statistical analysis. \\
& \textit{Codebook question}: None \\
\\
\multicolumn{2}{l}{\textbf{metric\_access}} \\
& \textit{Description}: Whether the metric requires model access or not. \\
& \textit{Codebook question}: Does this metric require model access, or can it be computed from responses alone? \\
\\
& \textit{Summary of values:} \\
& Outputs alone: 422 \\
& Model access required (e.g. logits): 32 \\
\\
\multicolumn{2}{l}{\textbf{metric\_definition\_detail}} \\
& \textit{Description}: Additional details on metric definition. \\
& \textit{Codebook question}: Any additional details on metric definition. (optional, only if something stood out) \\
\\
\multicolumn{2}{l}{\textbf{metric\_face\_validity}} \\
& \textit{Description}: An assessment of the face validity of the metric. \\
& \textit{Codebook question}: Is there prima facie reason to believe that this metric could benchmark the target phenomenon? \\
\\
\multicolumn{2}{l}{\textbf{metric\_face\_validity\_clean}} \\
& \textit{Description}: Standardised mapping of metric\_face\_validity values for statistical analysis. \\
& \textit{Codebook question}: None \\
\\
\multicolumn{2}{l}{\textbf{metric\_aggregation}} \\
& \textit{Description}: The method(s) used to aggregate metric scores. \\
& \textit{Codebook question}: How are the results aggregated, if at all? (e.g. mean, weighted mean, correlation) \\
\\
\multicolumn{2}{l}{\textbf{metric\_subscores}} \\
& \textit{Description}: Whether subscores are provided for any specific subsets of the task. \\
& \textit{Codebook question}: Are scores provided for any specific subsets of the task? (e.g. by difficulty) \\
\\
& \textit{Summary of values:} \\
& Yes: 363 \\
& No: 91 \\
\\
\multicolumn{2}{l}{\textbf{metric\_subscores\_detail}} \\
& \textit{Description}: Standardised mapping of metric\_subscores values for statistical analysis. \\
& \textit{Codebook question}: If so, what subsets are provided? \\
\\
\multicolumn{2}{l}{\textbf{metric\_metascoring}} \\
& \textit{Description}: Whether the scoring involves meta-scoring techniques (pass@k, consensus@k, etc.). \\
& \textit{Codebook question}: Does the scoring involve any meta-scoring techniques? If so, which ones? \\
\\
\multicolumn{2}{l}{\textbf{metric\_fewshot}} \\
& \textit{Description}: Whether evaluation uses few-shot prompting. \\
& \textit{Codebook question}: Does the scoring involve few-shot prompting or other similar in-context learning techniques? \\
\\
& \textit{Summary of values:} \\
& No: 214 \\
& Yes: 79 \\
\\
\multicolumn{2}{l}{\textbf{metric\_statistics}} \\
& \textit{Description}: The statistics used to aggregate and compare metric scores. \\
& \textit{Codebook question}: What statistical methods are used to aggregate and compare the results? (e.g. simple mean/sum, mean and variance, clustered standard deviations) \\
\\
\multicolumn{2}{l}{\textbf{metric\_statistics\_clean}} \\
& \textit{Description}: Standardised mapping of metric\_statistics values for statistical analysis. \\
& \textit{Codebook question}: None \\
\\
\end{longtable}}

\subsection{Results and \textcolor{claims}{Claims}}
{\small\begin{longtable}{p{.005\textwidth}p{.995\textwidth}}
\multicolumn{2}{l}{\textbf{result\_interpretation}} \\
& \textit{Description}: Connection between claims and phenomenon definition. \\
& \textit{Codebook question}: Are the claims made in the results consistent with the scope of the definition being used? \\
\\
& \textit{Summary of values:} \\
& Yes: 435 \\
& No: 18 \\
\\
\multicolumn{2}{l}{\textbf{results\_comparison}} \\
& \textit{Description}: Comparison of results to other benchmarks. \\
& \textit{Codebook question}: Are comparisons made to results on other benchmarks of similar phenomena? (this requires a comparison of the nature of the results, not just a literature review) \\
\\
& \textit{Summary of values:} \\
& No: 294 \\
& Yes: 160 \\
\\
\multicolumn{2}{l}{\textbf{results\_comparison\_explanation}} \\
& \textit{Description}: Free-form explanation of the comparison. \\
& \textit{Codebook question}: If so, are theories offered to explain the similarities and differences? \\
\\
& \textit{Summary of values:} \\
& No comparisons made: 261 \\
& Yes: 144 \\
& No: 27 \\
\\
\multicolumn{2}{l}{\textbf{results\_human\_baseline}} \\
& \textit{Description}: Whether model results are compared to human performance. \\
& \textit{Codebook question}: Does the paper present a human baseline on the task? \\
\\
& \textit{Summary of values:} \\
& No: 305 \\
& Yes: 146 \\
\\
\multicolumn{2}{l}{\textbf{results\_author\_validity}} \\
& \textit{Description}: Whether benchmark authors directly address construct validity. \\
& \textit{Codebook question}: Do the authors present their own assessment of the validity of their benchmark? (i.e. do they directly address the question of construct validity for their benchmark?) \\
\\
\multicolumn{2}{l}{\textbf{results\_author\_validity\_clean}} \\
& \textit{Description}: Standardised mapping of results\_author\_validity values for statistical analysis. \\
& \textit{Codebook question}: None \\
\\
\multicolumn{2}{l}{\textbf{results\_author\_validity\_detail}} \\
& \textit{Description}: Free-form explanation of results\_author\_validity. \\
& \textit{Codebook question}: If so, please describe their evidence. \\
\\
\multicolumn{2}{l}{\textbf{results\_realism}} \\
& \textit{Description}: Whether benchmark results compare to real settings. \\
& \textit{Codebook question}: Are comparisons made between the benchmark results and results from more realistic settings? (e.g. MedQA vs supporting doctors in practice) \\
\\
& \textit{Summary of values:} \\
& No: 308 \\
& The benchmark is itself realistic: 119 \\
& Yes: 23 \\
& Other: 4 \\
\\
\multicolumn{2}{l}{\textbf{results\_realism\_clean}} \\
& \textit{Description}: Standardised mapping of results\_realism values for statistical analysis. \\
& \textit{Codebook question}: None \\
\\
\end{longtable}}

\subsection{Procedural}
{\small\begin{longtable}{p{.005\textwidth}p{.995\textwidth}}
\multicolumn{2}{l}{\textbf{authorship}} \\
& \textit{Description}: Authorship composition (industry or academia). \\
& \textit{Codebook question}: Authorship composition of the article \\
\\
& \textit{Summary of values:} \\
& Academia: 230 \\
& Mix (multiple authors from industry and academia): 186 \\
& Industry: 33 \\
& Other: 4 \\
\\
\multicolumn{2}{l}{\textbf{benchmark\_availability}} \\
& \textit{Description}: Benchmark online availability. \\
& \textit{Codebook question}: Whether the benchmark artefact is publicly available \\
\\
\multicolumn{2}{l}{\textbf{benchmark\_location}} \\
& \textit{Description}: Benchmark URL. \\
& \textit{Codebook question}: A link to the benchmark, if available (GitHub or similar) \\
\\
\multicolumn{2}{l}{\textbf{procedural\_extra}} \\
& \textit{Description}: Optional additional notes on procedural details. \\
& \textit{Codebook question}: Other useful notes on procedural details  (optional, only if something stood out) \\
\\
\multicolumn{2}{l}{\textbf{notes\_extra}} \\
& \textit{Description}: Final optional notes. \\
& \textit{Codebook question}: Any final notes about the paper not covered by above sections \\
\\
\end{longtable}}

\clearpage
\section{Inclusion and Exclusion Process}
\label{app:schema}

\begin{figure}[t!]
    \centering
    \includegraphics[width=\linewidth]{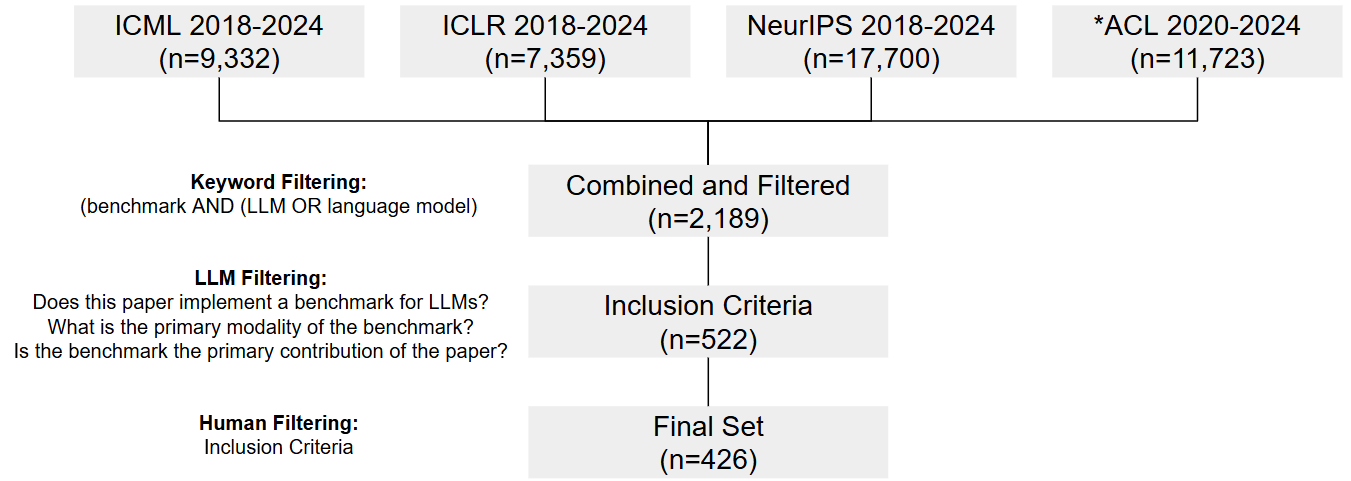}
    \caption{\textbf{Flowchart of the systematic review process.} Searching across EMNLP, NAACL, ACL, ICML, ICLR, and NeurIPS, we identified 2,189 papers matching the keyword search, and 445 which ultimately met the review criteria.}
    \label{fig:schema}
\end{figure}

We conducted a systematic review, using a combination of keyword search, LLM filtering, and human filtering to identify articles. Figure~\ref{fig:schema} shows the steps of the search process and the number of papers included at each step.

\subsection{Keyword Search and LLM Filtering}

Keyword search across the six target conferences identified 2,189 papers which included the keywords `benchmark' and `LLM' or `language model' in the title or abstract. A manual scan of these articles indicated that many articles were technical papers developing techniques for language modelling which reported improvements on various benchmarks. Since these papers were not the target of the review, further filtering was conducted via LLM.

For the LLM filtering, we used GPT-4o mini to identify the features about the papers relevant to inclusion and exclusion. We processed the inclusion criteria in order, removing articles at each step which did not meet the criteria. Table~\ref{tab:llm_exclusion} shows the steps of this process.

\begin{table}[h!]
    \centering
    \renewcommand{\arraystretch}{1.3}
    {\small
    \begin{tabular}{p{.2\textwidth}p{.45\textwidth}p{.1\textwidth}p{.1\textwidth}}
            \toprule
         \textbf{\newline Criterion} & \textbf{\newline LLM Prompt} & \textbf{Articles\newline Excluded} & \textbf{Articles\newline Remaining} \\
         \midrule
         System Prompt & You are an academic assistant, filtering articles to identify which ones are relevant to a literature review. & - & 2,189\\
         Exclude articles which do not implement a benchmark & Please read the paper title and abstract below, and tell me whether the paper creates and describes a new benchmark for large language models.
                  
         After reading the title and abstract, please very briefly describe whether the article implements a new benchmark for large language models. Then, on a new line write **Answer:** followed by a single word answer of `Yes' or `No' as to whether the article creates and describes a new benchmark. & 1,251 & 938 \\
         Exclude articles which require modalities beyond text and vision & Please read the paper title and abstract below, and tell me the primary modality of the dataset being used.
                  
        After reading the title and abstract, please very briefly describe the primary modality of the article. Then, on a new line write **Answer:** followed by a single word answer describing the primary modality considered in the article. Your answer should be either Language, Image, Video, Audio, Multimodal or Other. Use Other only when the primary modality is not one of the previous options. & 92 & 846 \\
        Exclude articles which are not primarily about creating a new benchmark & Please read the paper title and abstract below, and tell me the primary focus area of the paper.
                  
        After reading the title and abstract, please very briefly describe the primary focus of the paper. Then, on a new line write **Answer:** followed by a single word answer of `Benchmark', `Technical', `Methodological' or `Other' to categorize the primary contribution. & 324 & 522 \\
        \bottomrule
        \\
    \end{tabular}
    \caption{\textbf{LLM Filtering Steps.} The prompts used for progressive filtering of the articles to be reviewed, and the number of articles excluded at each step.}
    \label{tab:llm_exclusion}
    }
\end{table}

We validated the results of this process by randomly selecting 50 articles from the 2,189 and manually categorising them for inclusion and exclusion. Table~\ref{tab:exclusion_validation} shows the confusion matrix of the LLM exclusion relative to the human gold standard classes. The overall precision in this subset is 80\% and the recall is 89\%. As such we expect the filtering to be highly effective in capturing the relevant articles for human review, though not perfect, which we note as a limitation. 

Of the overall batch of 2,189, 522 articles were selected for review, about 24\%, which is similar to the 20\% of articles selected in the random sample. The fraction of the manually reviewed articles which were included in the final study was 445, indicating that the precision in the full study was about 85\%, similar to this sample. 

\begin{table}[h!]
    \centering
    \renewcommand{\arraystretch}{1.3}
{\small
    \begin{tabular}
    {p{.3\textwidth}p{.1\textwidth}p{.1\textwidth}p{.1\textwidth}p{.1\textwidth}p{.1\textwidth}}
    \toprule
        \textbf{\newline Criterion} & \textbf{Articles \newline Compared} & \textbf{True\newline Inclusion} & \textbf{False\newline Inclusion} & \textbf{True\newline Exclusion} & \textbf{False\newline Exclusion}\\
        \midrule
        Exclude articles which do not implement a benchmark & 50 & 14 & 9 & 22 & 5\\
        Exclude articles which require modalities beyond text and vision & 14 & 13 & 0 & 1 & 0\\
        Exclude articles which are not primarily about creating a new benchmark. & 9 & 8 & 0 & 0 & 1\\
        Overall & 50 & 8 & 2 & 39 & 1 \\ 
        \bottomrule
        \\
    \end{tabular}
    \caption{\textbf{LLM Filtering Human Validation.} A comparison between human and LLM filtering on a subset of 50 articles drawn at random from the initial list of 2,189. Human filtering results are treated as gold-standard. The steps were conducted in the same order as the real filtering, so that the number of papers remaining falls at each step. The confusion matrix is shown for each step of the filtering process.}
    \label{tab:exclusion_validation}
    }
\end{table}

\clearpage
\section{Inter-rater Agreement} \label{app:validation}

To assess the reliability of our annotation procedure and the consistency of coding judgments across reviewers, we measured inter-rater agreement on a randomly selected subset of 46 benchmark papers. Each paper in this subset was independently annotated by two reviewers using our codebook of 30 categorical items relating to phenomena, tasks, metrics, and validity claims.

Inter-rater agreement is essential in systematic reviews where subjectivity or interpretation may influence labelling decisions. High agreement indicates that the annotation schema is sufficiently well-defined and that results can be considered reliable across different coders. Conversely, low agreement may indicate ambiguity in the coding process. Prior work in empirical machine learning has emphasized the importance of such reliability assessments when coding qualitative features or design properties \cite{artstein2008inter, geiger2020garbage}.

Given the mix of binary, multi-class, and multi-label questions in our codebook, and the presence of strong label imbalance in many fields, we report agreement using both \textit{Percent Agreement} and \textit{Brennan–Prediger Kappa (BPK)} \cite{bpkappa}.

Percent agreement is computed as the proportion of items on which both raters agreed. For binary and multi-class fields, agreement is determined via exact label match. For multi-label questions (i.e. `check all that apply'), we compute the Jaccard similarity between the two sets of selected options for each item and take the average of these values across all items as the percent agreement score.

BPK adjusts for chance agreement under the assumption of uniform response distributions and is more robust to label imbalance than standard chance-corrected metrics such as Cohen’s Kappa, Fleiss’ Kappa, or Krippendorff’s Alpha, which often degrade in skewed settings. It is defined as:
\[
\text{BPK} = \dfrac{P_o - k^{-1}}{1 - k^{-1}},
\]
where \( P_o \) is the observed proportion of agreement and \( k \) is the number of valid response categories for the question. For binary fields (\(k = 2\)), this simplifies to $\text{BPK} = 2P_o - 1$. 

We compute BPK for all binary and multi-class single-label questions using the corresponding value of \(k\). For multi-label fields, we threshold the Jaccard similarity at 0.3 and treat item pairs with similarity above this threshold as “agreed”, thus converting the comparison into a binary decision before applying BPK. While BPK does not handle missing values, Krippendorff’s Alpha does, it is less suited to class-imbalanced data, and therefore we prefer BPK in our setting.

Across the 30 annotated fields, we observe a mean percent agreement of $68.1\%$ and a mean BPK of $0.524$, indicating overall moderate consistency. Structured and objective fields  exhibit very high agreement. In contrast, interpretive or compositional fields such as \texttt{task\_ecology} and \texttt{dataset\_sampling\_method} show lower consistency, highlighting areas where definitions could be improved.

\begin{table}[h!]
\centering
\caption{Inter-rater agreement across all fields.}
\label{tab:irr-results}
\begin{tabular}{lccc}
\toprule
\textbf{Codebook Field} & \textbf{Question Type} & \textbf{Percent Agreement (\%)} & \textbf{BPK} \\
\midrule
task\_face\_validity          & Multi-class     & 95.12 & 0.927 \\
metric\_access                & Binary          & 95.12 & 0.902 \\
benchmark\_availability       & Multi-class     & 95.12 & 0.939 \\
inclusion                     & Binary          & 93.48 & 0.870 \\
result\_interpretation        & Binary          & 92.68 & 0.854 \\
metric\_aggregation           & Multi-label     & 87.20 & 0.756 \\
metric\_face\_validity        & Multi-class     & 85.37 & 0.780 \\
task\_train\_val              & Multi-label     & 84.15 & 0.951 \\
metric\_metascoring           & Multi-class     & 82.93 & 0.787 \\
results\_human\_baseline      & Binary          & 80.49 & 0.610 \\
task\_dataset\_metadata       & Binary          & 75.61 & 0.512 \\
metric\_fewshot               & Binary          & 73.17 & 0.463 \\
authorship                    & Multi-class     & 73.17 & 0.642 \\
response\_format              & Multi-label     & 71.54 & 0.707 \\
metric\_subscores             & Binary          & 70.73 & 0.415 \\
definition\_integrity\_detail & Multi-class     & 60.98 & 0.415 \\
results\_comparison           & Binary          & 60.98 & 0.220 \\
phenomenon\_short             & Multi-label     & 60.98 & 0.220 \\
results\_realism              & Multi-class     & 58.54 & 0.447 \\
definition\_integrity         & Multi-class     & 58.54 & 0.378 \\
definition\_scope             & Multi-class     & 56.10 & 0.341 \\
results\_comparison\_explanation & Multi-class & 56.10 & 0.341 \\
metric\_definition            & Multi-label     & 55.37 & 0.512 \\
task\_source                  & Multi-label     & 54.23 & 0.561 \\
results\_author\_validity     & Multi-class     & 53.66 & 0.305 \\
phenomenon\_defined           & Binary          & 53.66 & 0.073 \\
phenomenon\_contested         & Multi-class     & 48.78 & 0.317 \\
exclusion\_criteria           & Multi-class     & 40.00 & 0.200 \\
dataset\_sampling\_method     & Multi-label     & 37.80 & 0.122 \\
task\_ecology                 & Multi-class     & 31.71 & 0.146 \\
\midrule
\textbf{Mean}                 & —               & \textbf{68.11} & \textbf{0.524} \\
\bottomrule
\end{tabular}
\end{table}

These findings support the overall reliability of our coding process and offer a basis for targeted improvements in future annotation protocols, especially in interpretive or multi-choice fields.

% \clearpage

\printbibliography[title={Additional Reviewed Papers}]
\end{refsection}
\clearpage

\end{document}